\documentclass[conference]{IEEEtran}
\IEEEoverridecommandlockouts

\usepackage{cite}
\usepackage{amsmath,amssymb,amsfonts}
\usepackage{graphicx}
\usepackage{textcomp}
\usepackage{xcolor}
\usepackage{booktabs}
\usepackage{array}
\usepackage{multirow}
\usepackage{dblfloatfix}

\begin{document}

\title{Camera-Noise Residuals for Face-Swap Detection:\\
Redundant, Not Complementary, and Why}

\author{%
\IEEEauthorblockN{Danil Davydov\IEEEauthorrefmark{1},
Bader Rasheed\IEEEauthorrefmark{2},
Dmitriy Vatolin\IEEEauthorrefmark{2}}
\IEEEauthorblockA{%
\IEEEauthorrefmark{1}\textit{Innopolis University}, Innopolis, Russia\\
\IEEEauthorrefmark{2}\textit{Laboratory of Innovative Technologies for Processing
Video Content, Innopolis University}, Innopolis, Russia\\
\texttt{\{d.davydov, b.rasheed, d.vatolin\}@innopolis.university}}}

\maketitle

\begin{abstract}
Fusing a learned camera-noise fingerprint with an RGB appearance backbone is an
appealing route to generator-independent deepfake detection, because the noise
residual is grounded in image-formation physics rather than in the texture
statistics of a particular generator. We test, on FaceForensics++, whether a
Noiseprint++ residual channel carries information \emph{complementary} to an RGB
Xception backbone for face-swap detection. A three-model ablation (RGB-only,
residual-only, late-fusion) shows that fusion does not improve over RGB alone and
that the residual branch alone is near chance. A seven-level bottleneck diagnostic
localizes the cause: the noise maps do carry a discriminative signal, but it is
statistical---carried by the per-sample first and second moments (mean, variance,
energy) of the residual---and the per-sample \texttt{InstanceNorm} layer placed at
the noise-branch input, following the TruFor template, standardizes exactly those
moments away (five-fold cross-validated AUC drops from $0.747$ to $0.554$). A
context-crop control rules out cropping geometry, and two fixed-fusion variants that
remove the bottleneck recover the statistical signal yet still fail to beat RGB on
every dataset. We conclude that, on this manipulation distribution, the noise
residual is redundant with RGB rather than complementary, and we give concrete
guidance for practitioners adopting noise-residual fusion for face-swap detection.
\end{abstract}


\section{Introduction}
Face-swap deepfakes have become photorealistic and cheap to produce, eroding the
reliability of visual media~\cite{rossler2019faceforensics,dolhansky2020dfdc}.
Detectors trained on the texture-level artefacts of one generator family often fail
when confronted with a new architecture or dataset, behaving as fingerprint matchers
that memorize dataset-specific quirks rather than validators of physical
plausibility. This motivates detection signals grounded in invariants of image
formation, which a generator would have to reproduce by implicitly solving the
underlying inverse problems.

The sensor-noise residual is the canonical such invariant. Photo-response
non-uniformity and learned camera-model fingerprints survive throughout a device's
lifetime and are disrupted where two images are composited~\cite{cozzolino2020noiseprint}.
Noiseprint~\cite{cozzolino2020noiseprint} extracts a camera-model fingerprint
disentangled from scene content, and TruFor~\cite{guillaro2023trufor} builds an
end-to-end forgery detector around the Noiseprint++ fingerprint with a learned fusion
head. These methods are reported as effective in general image-forgery settings, but
the specific question of whether the noise channel is \emph{complementary} to an RGB
appearance backbone \emph{on face-swap manipulations}---as opposed to redundant with
it, or destroyed by architectural choices---has not been settled by a controlled
ablation.

\textbf{Terminology.} We use \emph{complementary} and \emph{redundant} in a precise,
measurable sense throughout. Let $A(\cdot)$ denote detection AUC for a given channel
or fusion of channels. Two channels are \textbf{complementary} when fusing them
strictly exceeds the better single channel beyond statistical noise,
\[
A(\text{fusion}) > \max\{A(\text{RGB}),\,A(\text{residual})\},
\]
i.e.\ the second channel supplies discriminative information the first does not. They
are \textbf{redundant} when fusion fails to exceed the stronger single channel,
\[
A(\text{fusion}) \approx \max\{A(\text{RGB}),\,A(\text{residual})\} \le A(\text{RGB}),
\]
i.e.\ the second channel adds nothing the first has not already captured. Our question
is whether the Noiseprint++ residual is complementary to RGB in this sense for
face-swap detection.

We address that question directly on FaceForensics++ (FF++)~\cite{rossler2019faceforensics}.
Our contributions are:
\begin{itemize}
\item A controlled three-model ablation (RGB-only, residual-only, late-fusion)
showing that an architecturally fused Noiseprint++ residual channel does
\emph{not} improve over RGB alone on FF++ (Wilcoxon $p=0.977$), while the
residual branch alone is near chance.
\item A seven-level bottleneck diagnostic that localizes the failure of the
residual branch to the per-sample \texttt{InstanceNorm} input layer: a classifier
on six scalar noise statistics reaches AUC $0.747$, but the same statistics after
\texttt{InstanceNorm} pooling collapse to $0.554$.
\item A context-crop control and two five-seed fixed-fusion variants that rule out
cropping geometry and the normalization bottleneck as the cause of the fusion
failure, leaving redundancy with the RGB channel as the explanation supported by
the evidence---together with practitioner guidance.
\end{itemize}

\section{Related Work}
\textbf{Deepfake detection and generalization.}
FF++~\cite{rossler2019faceforensics} established that convolutional detectors achieve
strong in-distribution accuracy but degrade sharply across manipulation methods and
datasets. The Self-Blended Images (SBI) framework~\cite{shiohara2022sbi}, building on
Face X-Ray~\cite{li2020facexray}, generates training fakes by self-blending a single
source image, removing the dependence on labelled fakes and yielding markedly better
cross-dataset transfer; we use an off-the-shelf SBI model as a reference baseline.

\textbf{Sensor-noise forensics.}
Noiseprint~\cite{cozzolino2020noiseprint} learns a camera-model fingerprint via a
Siamese contrastive objective on same-camera versus different-camera patches.
TruFor~\cite{guillaro2023trufor} couples the Noiseprint++ fingerprint with an RGB
stream through a learned fusion head and places an \texttt{InstanceNorm} layer at the
noise-branch input. Our study adopts that architectural template as the basis for a
face-swap-specific ablation and asks where, in the extraction--processing--fusion
pipeline, the signal is or is not retained.

\section{Method}
\subsection{Data and noise extraction}
All experiments use face crops from FF++~\cite{rossler2019faceforensics} and its
DeepFakeDetection extension, partitioned \emph{at the video level} so that no source
video contributes frames to more than one split ($19{,}571$ train, $3{,}573$ val,
$4{,}675$ test crops). Faces are detected with MTCNN~\cite{zhang2016mtcnn}, the box
expanded by $1.3\times$ and square-padded. Noise maps are precomputed with
Noiseprint++~\cite{guillaro2023trufor} on the \emph{full frame at native resolution
before} cropping; this extract-then-crop order preserves the camera-model fingerprint
at the face location, whereas running the extractor on the small crop degrades its
receptive-field context. Class imbalance is handled with class-weighted
cross-entropy. The full data protocol and training configuration are tabulated in the
Appendix (Table~\ref{tab:protocol}).

\subsection{Three ablation models}
\begin{itemize}
\item \textbf{RGB-only:} an ImageNet-pretrained Xception~\cite{chollet2017xception}
backbone on the face crop with a three-layer head---isolating the RGB appearance
signal.
\item \textbf{Residual-only:} a ResNet~\cite{he2016resnet} backbone on the
$299\times299$ Noiseprint++ map with the identical head; its first layer is
\texttt{InstanceNorm2d}, following the TruFor architecture~\cite{guillaro2023trufor}.
\item \textbf{Late-Fusion:} parallel RGB and noise branches concatenated into a joint
head---testing whether the noise signal is complementary to RGB.
\end{itemize}
Models are trained with AdamW, a tri-rate learning-rate schedule with three-epoch
warmup, \texttt{ReduceLROnPlateau} on validation macro-F1, gradient clipping, and a
progressive backbone-unfreezing schedule.

\subsection{Seven-level bottleneck diagnostic}
To localize the residual branch's failure independently of the deep network, we run a
progressive probe on the precomputed noise maps via five-fold cross-validation on
$3{,}000$ FF++ test samples. The seven levels are: (L1) logistic regression on mean
$|\text{noise}|$; (L2) logistic regression on six scalar statistics (mean, std,
energy, max, kurtosis, pixel count); (L3) a small MLP on those six features; (L4)
logistic regression on \texttt{InstanceNorm}-pooled noise; (L5) logistic regression on
raw-pooled noise; (L6) a two-layer ConvNet on raw noise; (L7) the same ConvNet with
\texttt{InstanceNorm} before the first convolution. L1--L3 test whether scalar statistics suffice; L4 (standardize-then-pool) tests whether removing the per-sample moments collapses a linear readout, and L7 tests whether a learned convolution can recover signal after standardization; L5--L6 test raw spatial patterns.

\subsection{Context-crop control}
We additionally test whether expanding the crop from $1.3\times$ to $2.7\times$ the
box---to capture the blending boundary where PRNU discontinuities should be most
visible---helps. The same probe is applied to both crop scales on $2{,}000$ samples
each from FF++ and Celeb-DF, alongside a classifier-free corpus signal-to-noise ratio
(SNR) comparing between-class to within-class variance of the noise statistics.

\subsection{Fixed-fusion variants}
Finally, two variants (five seeds each) directly remove the diagnosed bottleneck:
\textbf{StatNoise-Fusion} bypasses spatial processing and fuses thirteen scalar noise
statistics with Xception features through a small MLP; \textbf{ResAware-Fusion}
replaces the noise branch's \texttt{InstanceNorm} with \texttt{BatchNorm} and processes
the noise crop at native resolution, preserving the per-sample statistics through the
spatial pipeline.

\section{Results}
\subsection{Ablation and cross-dataset transfer}
Table~\ref{tab:main} reports video-level AUC (frame-level for DFF)
in distribution and across three further corpora~\cite{li2020celebdf,dolhansky2020dfdc,song2023dff}.
RGB-only and Late-Fusion are statistically indistinguishable on FF++ (Wilcoxon paired
test on per-video squared errors, $p=0.977$ after Bonferroni correction), the
residual branch is near chance everywhere, and the SBI reference---an orthogonal
training-data prior---transfers best across datasets. Adding the noise branch buys
nothing over RGB: RGB-Only and Late-Fusion are indistinguishable at both frame
and video level, while Residual-Only stays near chance.

\begin{table}[t]
\caption{In-distribution and cross-dataset video-level AUC (frame-level for DFF).
RGB-only $\approx$ Late-Fusion (Wilcoxon $p=0.977$); Residual-only is near chance.
SBI is an off-the-shelf reference, shown for context.}
\label{tab:main}
\centering
\footnotesize
\setlength{\tabcolsep}{4pt}
\begin{tabular}{lrrrr}
\toprule
\textbf{Model} & \textbf{FF++} & \textbf{Celeb-DF} & \textbf{DFDC} & \textbf{DFF} \\
\midrule
RGB-Only      & $\mathbf{0.928}$ & $0.888$ & $0.717$ & $0.531$ \\
Late-Fusion   & $0.924$ & $0.881$ & $0.712$ & $0.475$ \\
Residual-Only & $0.579$ & $0.547$ & $0.578$ & $0.212$ \\
\textit{SBI (ref.)} & $0.865$ & $\mathbf{0.907}$ & $\mathbf{0.884}$ & $\mathbf{0.614}$ \\
\bottomrule
\end{tabular}
\end{table}

\subsection{The bottleneck is per-sample \texttt{InstanceNorm}}
Table~\ref{tab:bottleneck} resolves where the residual signal is lost. The noise maps
\emph{do} carry a discriminative signal, and it is statistical: an MLP on six scalar
statistics (L3) reaches AUC $0.747$, well above chance. But the same statistics after
\texttt{InstanceNorm} pooling (L4) collapse to $0.554$.

\begin{figure}[t]
\centering
\includegraphics[width=\columnwidth]{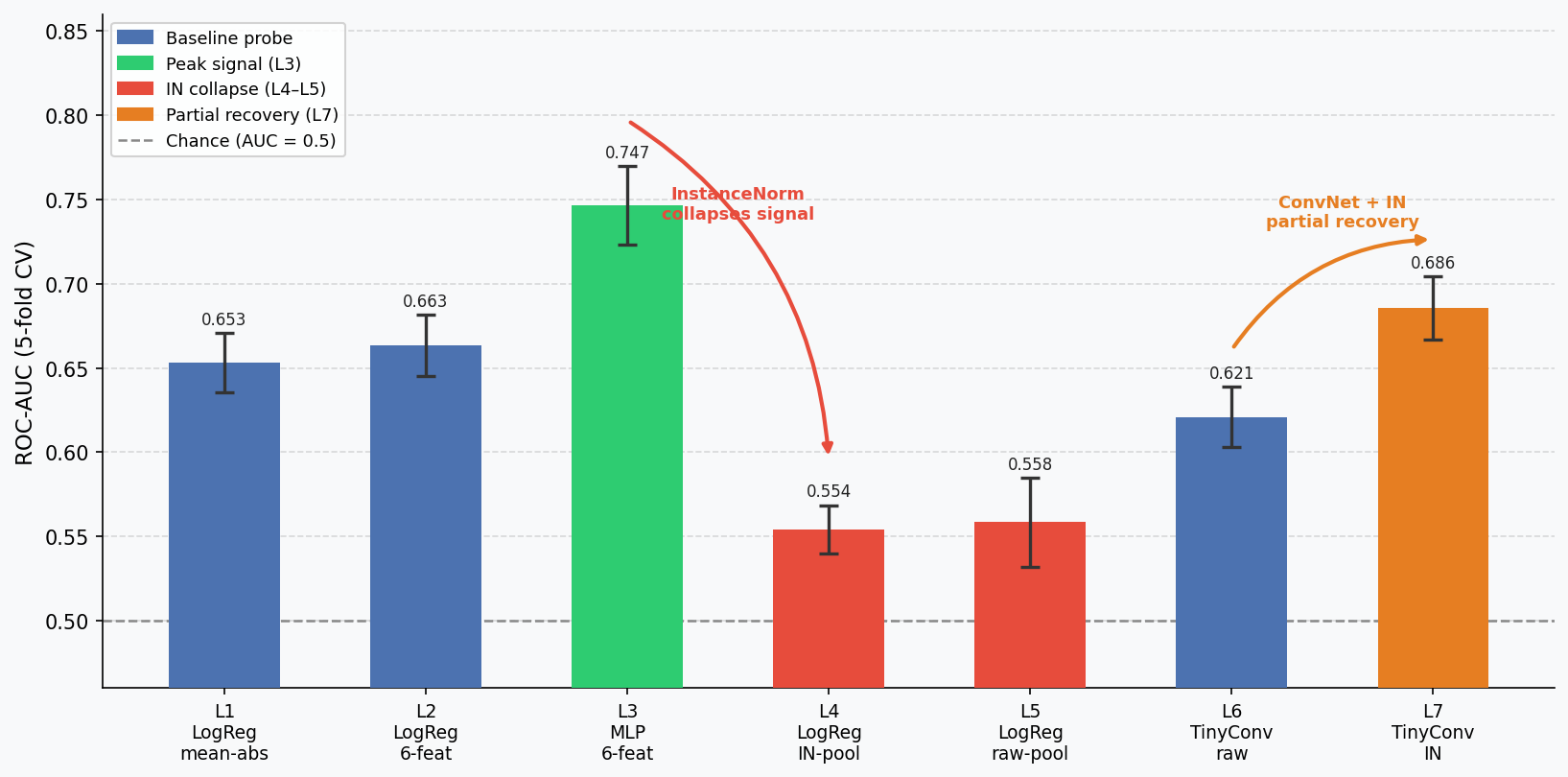}
\caption{Seven-level bottleneck AUC (L1--L7) with five-fold standard-deviation bars.
The L3 $\rightarrow$ L4 drop localizes the loss to standardize-then-pool; L7 recovers part of the statistical ceiling.}
\label{fig:bottleneck}
\end{figure}

To see why, consider the operation itself. For a single noise map
$x\in\mathbb{R}^{C\times H\times W}$, \texttt{InstanceNorm} standardizes each channel
$c$ independently using only that sample's own spatial statistics,
\begin{equation}
\hat{x}_{c,h,w} \;=\; \gamma_c\,
\frac{x_{c,h,w}-\mu_c}{\sqrt{\sigma_c^2+\epsilon}} \;+\; \beta_c,
\label{eq:in}
\end{equation}
\begin{equation}
\mu_c=\frac{1}{HW}\sum_{h,w} x_{c,h,w},
\qquad
\sigma_c^2=\frac{1}{HW}\sum_{h,w}\bigl(x_{c,h,w}-\mu_c\bigr)^2 .
\label{eq:in-stats}
\end{equation}
Eq.~\eqref{eq:in} subtracts the per-sample mean $\mu_c$ and divides by the per-sample
standard deviation $\sigma_c$. It therefore discards exactly the first- and
second-moment information of each noise map---the per-sample mean and the
variance/energy---and rescales any scale-dependent descriptor such as $\max|x|$ to a
common per-sample scale. These are precisely the quantities that dominate the
discriminative six-feature set (mean, std, energy, max). Standardized higher-order
moments such as kurtosis are invariant to the affine normalization in
Eq.~\eqref{eq:in} and so survive it in principle; the empirical
$\text{L3}\!\rightarrow\!\text{L4}$ collapse to near chance shows that, once the first
two moments are removed, the residual carries little discriminative signal on its own.
The Residual-Only ConvNet, which applies \texttt{InstanceNorm} before any deep
convolution, therefore never sees the discriminative statistics; L7 (ConvNet with
\texttt{InstanceNorm}) recovers only part of the ceiling.
Fig.~\ref{fig:bottleneck} visualizes the L3$\rightarrow$L4 collapse.

\begin{table}[t]
\caption{Seven-level bottleneck diagnostic on Noiseprint++ maps. Five-fold
cross-validated AUC on $3{,}000$ FF++ test samples. L3 sets the statistical ceiling
($0.747$); L4 localizes the \texttt{InstanceNorm} collapse ($0.554$).}
\label{tab:bottleneck}
\centering
\footnotesize
\begin{tabular}{llr}
\toprule
\textbf{Classifier} & \textbf{Input} & \textbf{AUC} \\
\midrule
L1: LogReg            & mean $|$noise$|$            & $0.653 \pm 0.018$ \\
L2: LogReg            & 6 scalar stats             & $0.664 \pm 0.018$ \\
L3: MLP (ceiling)     & 6 scalar stats             & $\mathbf{0.747 \pm 0.023}$ \\
L4: LogReg            & InstanceNorm-pooled noise  & $0.554 \pm 0.014$ \\
L5: LogReg            & raw-pooled noise           & $0.559 \pm 0.026$ \\
L6: ConvNet (2-layer) & raw noise                  & $0.602 \pm 0.021$ \\
L7: ConvNet + IN      & raw noise                  & $0.670 \pm 0.014$ \\
\bottomrule
\end{tabular}
\end{table}

\subsection{Cropping geometry is not the cause}
The tight crop outperforms the context crop at every probe level on both datasets. On
FF++, corpus SNR is $0.188$ (tight) versus $0.080$ (context) and best-classifier AUC
$0.670$ versus $0.652$; on Celeb-DF, SNR $0.158$ versus $0.053$ and AUC $0.588$ versus
$0.580$. Expanding the crop dilutes the face-region signal with background pixels that
are identical between real and fake, lowering rather than raising the SNR. The
expected PRNU discontinuity at the blending boundary does not, on these manipulations,
yield a stronger signal than the face region itself.

\subsection{Fixing the bottleneck does not recover complementarity}
Table~\ref{tab:fixed} reports the two fixed-fusion variants. Both deliver a non-trivial
noise contribution at the fusion point---confirming the diagnosis---yet both
\emph{underperform} RGB-only and the original Late-Fusion on every dataset. The
original Late-Fusion learned to ignore a near-chance noise gradient and tracked
RGB-only; the fixed variants receive a stronger but redundant noise signal and are
pulled down by the weaker modality. This is the signature of redundancy, not
complementarity.

\begin{table}[t]
\caption{Fixed-fusion variants (mean $\pm$ std over five seeds). Removing the
\texttt{InstanceNorm} bottleneck recovers the statistical signal but still
underperforms RGB-only on both datasets.}
\label{tab:fixed}
\centering
\footnotesize
\setlength{\tabcolsep}{6pt}
\begin{tabular}{lrr}
\toprule
\textbf{Model} & \textbf{FF++ AUC} & \textbf{Celeb-DF AUC} \\
\midrule
RGB-Only (ref.)            & $\mathbf{0.928}$ & $\mathbf{0.888}$ \\
Late-Fusion (original)     & $0.924$ & $0.881$ \\
Residual-Only (original)   & $0.579$ & $0.547$ \\
StatNoise-Fusion (fixed)   & $0.845 \pm 0.009$ & $0.786 \pm 0.015$ \\
ResAware-Fusion (fixed)    & $0.856 \pm 0.009$ & $0.805 \pm 0.015$ \\
\bottomrule
\end{tabular}
\end{table}

\subsection{Robustness corroborates redundancy}
\begin{figure*}[!tb]
\centering
\includegraphics[width=\textwidth]{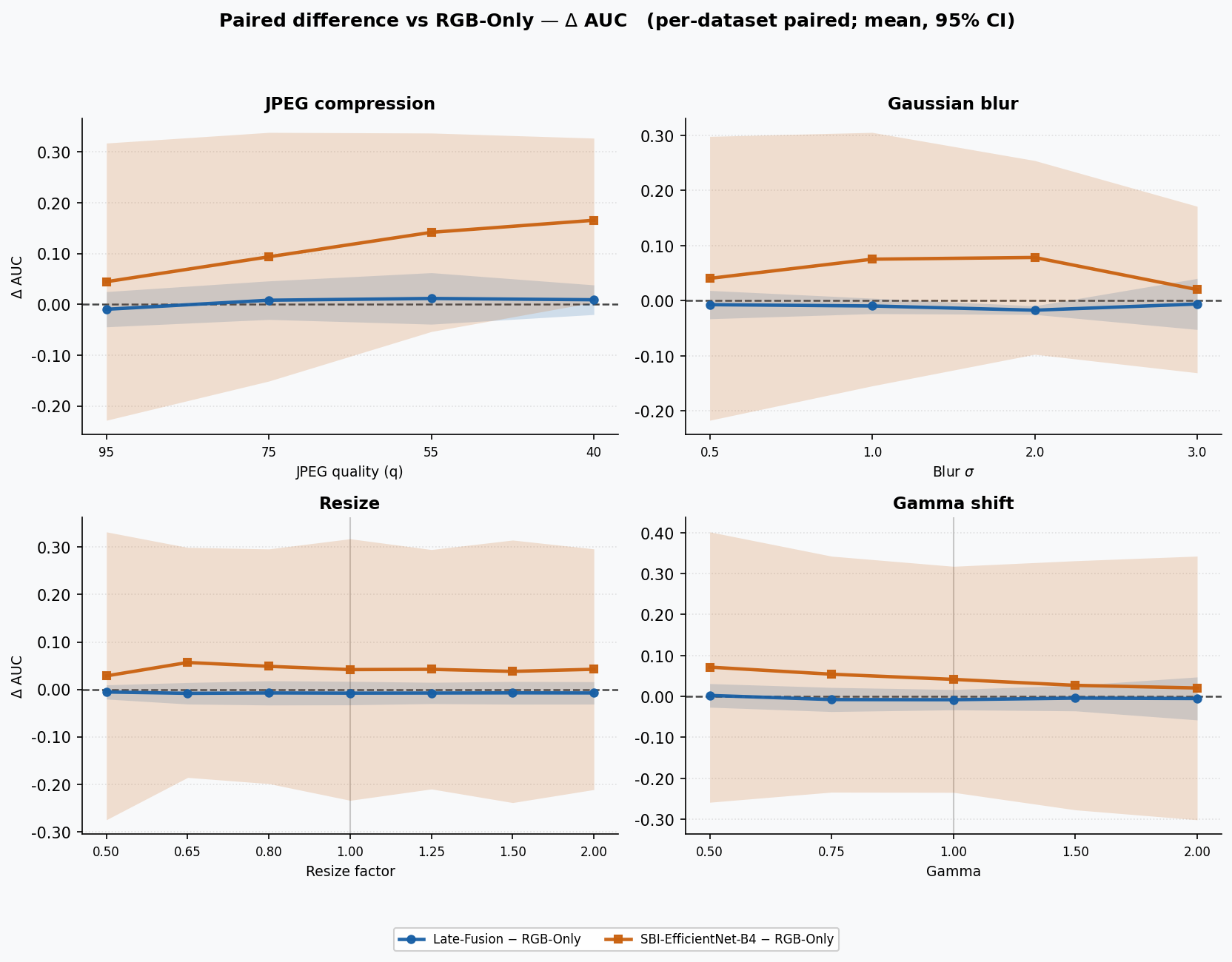}
\caption{Paired per-dataset difference in frame-level AUC relative to RGB-Only, by
perturbation family (JPEG, Gaussian blur, resize, gamma), across FF++, Celeb-DF, and
DFDC; shaded bands are $95\%$ CIs ($n=3$ datasets, paired within dataset). \textbf{Late-Fusion
$-$ RGB-Only} (blue) hugs the zero line at every operating point---never significantly
positive, its only CI-excluding-zero point being negative (blur $\sigma=2$)---so the
noise branch adds nothing over RGB. \textbf{SBI $-$ RGB-Only} (orange), included as a
reference for what a genuinely different model looks like, is positive throughout and
largest under heavy JPEG. DFF is excluded (image-only and near chance for the RGB and
fusion models).}
\label{fig:robust}
\end{figure*}

A genuinely complementary noise channel should make Late-Fusion behave differently
from RGB-Only under perturbations that degrade texture and sensor noise at different
rates. To test this directly we compute, for each perturbation family, the
\emph{paired} per-dataset difference $\Delta\mathrm{AUC}=\text{Late-Fusion}-\text{RGB-Only}$
at frame level across the three video datasets (FF++, Celeb-DF, DFDC; DFF is excluded
as image-only and near chance for both models), pairing within each dataset so that
between-dataset scale differences cancel. Fig.~\ref{fig:robust} plots $\Delta\mathrm{AUC}$
with $95\%$ confidence intervals.

The fusion difference is statistically indistinguishable from zero across the entire
sweep. Pooled over all operating points the mean $\Delta\mathrm{AUC}$ is $-0.003$
(max $|\Delta|=0.035$); it is never significantly positive at any JPEG, blur, resize,
or gamma level, and the only point whose CI excludes zero is \emph{negative} (Gaussian
blur $\sigma=2$: $-0.017$). Adding the noise branch therefore never improves robustness
and, where it has any measurable effect, slightly degrades it---the paired-perturbation
counterpart of the in-distribution Wilcoxon result ($p=0.977$). For contrast, the same
paired difference for the SBI reference ($\text{SBI}-\text{RGB-Only}$) is positive at
every operating point (means $+0.04$ to $+0.17$, largest under heavy JPEG), confirming
that the comparison readily resolves a genuinely different model. The noise-fusion curve,
by contrast, sits on the zero line---the most direct evidence that the fusion head has
learned to ignore the near-chance noise gradient rather than exploit it.

\section{Discussion}
The evidence supports a single, mechanistic account. (i) The Noiseprint++ residual
\emph{does} carry a face-swap signal on FF++, but it is statistical---differences in
the per-sample first and second moments (mean, variance/energy) of the
residual---consistent with the Noiseprint family's premise that compositing disrupts
the per-region camera-model fingerprint~\cite{cozzolino2020noiseprint}. (ii) That
signal is destroyed inside the trained branch by the per-sample \texttt{InstanceNorm}
layer that the TruFor template places at the input~\cite{guillaro2023trufor}: as
Eq.~\eqref{eq:in} shows, standardizing each sample to zero mean and unit variance
removes precisely the first two moments that carry the signal. The collapse is
invisible in loss curves and is only exposed by the classifier-free probe. (iii) Once
the bottleneck is removed, the recovered signal is \emph{redundant} with what the RGB
backbone already extracts from the same blending operation: the fixed-fusion variants
never exceed RGB-only, and the original Late-Fusion tracks RGB-only across the
perturbation sweep (Fig.~\ref{fig:robust}), consistent with the fusion head learning to
ignore the noise gradient. (iv) Cropping geometry is not responsible---the tight crop is
uniformly better.

The practical guidance follows directly. A noise-residual side channel adopted from a
general image-forgery detector should not be assumed complementary for face-swap
detection: if the discriminative signal is statistical, a per-sample normalization at
the branch input will erase it, and even when preserved it may add nothing over RGB.
The contrast with the SBI reference is instructive: a physical prior injected at the
\emph{training-data} level transfers across datasets far better than this architectural
side channel, suggesting that the locus of injection matters more than the presence of
the prior.

\textbf{Limitations.} The study is on the FF++ face-swap distribution; the conclusion
is specific to that manipulation family and to the Noiseprint++/TruFor template. We do
not claim sensor-noise residuals are useless for forgery detection in general---only
that they are redundant, not complementary, for face-swap detection here, and that a
common architectural choice silently destroys the signal they do carry.

\section{Conclusion}
We asked whether a learned camera-noise residual is complementary to an RGB backbone
for face-swap detection, and---if not---where the signal is lost. On FaceForensics++,
architectural fusion gives no improvement over RGB; a seven-level probe localizes the
failure to a per-sample \texttt{InstanceNorm} layer that erases the statistical signal
the residual carries (AUC $0.747\!\rightarrow\!0.554$); a context-crop control rules out
cropping geometry; and two fixed-fusion variants show that, once the signal is
preserved, it is redundant with RGB. The bottleneck probe is a reusable diagnostic for
auditing where an auxiliary forensic channel succeeds or fails.

\appendices
\section{Data Protocol and Training Details}
\label{app:protocol}
Table~\ref{tab:protocol} consolidates the dataset partitioning, preprocessing, and
training configuration used for all three ablation models, for reproducibility.

\begin{table*}[!tb]
\caption{Data protocol and training configuration. Splits are at the video level so
no source video crosses partitions.}
\label{tab:protocol}
\centering
\footnotesize
\renewcommand{\arraystretch}{1.15}
\begin{minipage}[t]{0.48\textwidth}
\centering
\begin{tabular}{@{}p{0.34\linewidth} p{0.60\linewidth}@{}}
\toprule
\textbf{Item} & \textbf{Setting} \\
\midrule
\multicolumn{2}{@{}l}{\textit{Data}}\\
Source corpus       & FF++ + DeepFakeDetection extension \\
Split granularity   & Video level (no leakage across splits) \\
Train / val / test  & $19{,}571$ / $3{,}573$ / $4{,}675$ crops \\
Cross-dataset eval  & Celeb-DF, DFDC, DFF \\
Class balance       & Class-weighted cross-entropy \\
\midrule
\multicolumn{2}{@{}l}{\textit{Preprocessing}}\\
Face detection      & MTCNN \\
Crop box            & $1.3\times$ expansion, square-padded \\
Input resolution    & $299\times299$ \\
Noise extractor     & Noiseprint++ (TruFor) \\
Extraction order    & Full frame at native res., then crop \\
\bottomrule
\end{tabular}
\end{minipage}
\hfill
\begin{minipage}[t]{0.48\textwidth}
\centering
\begin{tabular}{@{}p{0.36\linewidth} p{0.58\linewidth}@{}}
\toprule
\textbf{Item} & \textbf{Setting} \\
\midrule
\multicolumn{2}{@{}l}{\textit{Architectures}}\\
RGB branch          & Xception (ImageNet-pretrained) + 3-layer head \\
Residual branch     & ResNet + 3-layer head; \texttt{InstanceNorm2d} input \\
Late-Fusion         & Concatenated RGB + noise branches, joint head \\
\midrule
\multicolumn{2}{@{}l}{\textit{Training}}\\
Optimizer           & AdamW \\
LR schedule         & Tri-rate, 3-epoch warmup \\
LR plateau control  & \texttt{ReduceLROnPlateau} on val macro-F1 \\
Backbone unfreezing & Progressive \\
Stabilization       & Gradient clipping \\
Loss                & Class-weighted cross-entropy \\
\midrule
\multicolumn{2}{@{}l}{\textit{Evaluation}}\\
Primary metric      & ROC-AUC \\
Aggregation         & Video-level (mean); frame-level for DFF \\
Bottleneck probe    & 5-fold CV, $3{,}000$ FF++ test samples \\
Fixed-fusion variants & 5 seeds each \\
\bottomrule
\end{tabular}
\end{minipage}
\end{table*}


\end{document}